\documentclass{article}
\usepackage{spconf,amsmath,amssymb,booktabs,multirow,graphicx,hyperref}
\newcommand{\shrink}{\operatorname{shrink}}
\newcommand{\slerp}{\operatorname{slerp}}
\title{BINDCLIP: ONE BALANCED COUPLING FOR COMPOSITIONAL VISION--LANGUAGE SCORING}
\name{\begin{tabular}{@{}c@{}}
Liuyang Song$^{1}$ \quad Yi Zhang$^{2}$ \quad Zhongyi Deng$^{3}$ \quad
Daqian Yang$^{1}$ \quad Hongbo Zhang$^{1,\dagger}$
\end{tabular}}
\address{$^{1}$Peking University \quad $^{2}$Dongguan University of Technology \quad $^{3}$Sichuan Agricultural University}
\begin{document}
\ninept
\raggedbottom
\maketitle
\begingroup\let\thefootnote\relax
\footnotetext{\raggedright $^{\dagger}$Corresponding author: \texttt{hobezhang816@gmail.com}\par}
\endgroup

\begin{abstract}
Global vision--language similarities compress an image and a caption into one vector, preserving semantics but not which word corresponds to which region or how those regions are arranged; a model can recognize every word and object yet prefer a compositionally incorrect caption. We argue that a frozen encoder retains this association structure, so the problem is to read it rather than to rebuild it beside the pretrained similarity. We introduce BindCLIP, a pairwise scorer built on one latent object: a balanced token--patch--depth optimal-transport coupling that places both candidate captions and several visual depths in a single plan. Semantic, entity, order, and spatial evidence are read as energies of this state, and exchanging the candidates permutes the plan, making the score exactly antisymmetric. A geometric refinement inside the coupling contracts moves that the candidates and the visual depths do not support. No task label, parser, relation inventory, or detector is used. One checkpoint and one inference path improve the official What'sUp, ARO, and SugarCrepe benchmarks over frozen global CLIP, with the strongest transfer on the relation splits. Controls rule out patch access and caption-length shortcuts, and an inference-time lesion localizes spatial arrangement to the coupling.
\end{abstract}


\section{Introduction}
CLIP-style dual encoders reduce an image and a caption to one vector each and compare them with a single inner product, which makes large-scale retrieval and zero-shot transfer practical \cite{clip,siglip}. That compression need not preserve which attribute modifies which object, which object fills which role, or whether one region lies above another, so a model may recognize every relevant word and object yet still prefer a compositionally incorrect caption. Benchmarks make the failure precise: ARO shows that strong retrieval models behave like bags of words \cite{aro}, SugarCrepe removes the language-only shortcuts that templated negatives leak \cite{sugarcrepe}, later suites tighten the negatives against modern models and against purely lexical variation \cite{conme,sc2}, and What'sUp varies elementary spatial configurations while holding object identity nearly fixed \cite{whatsup}; surveys treat the gap as structural rather than incidental \cite{cvrsurvey}. All of them motivate pairwise scoring between plausible captions for one image.

A common repair builds additional structure beside the pretrained similarity: hard-negative finetuning for order sensitivity \cite{aro}, scene-graph parsing bound to object slots \cite{occlip}, explicit structural correspondence \cite{structclip}, synthetic perturbations that supervise the exact point of change \cite{sparcl}, or, without any training, recomposition over parser-derived subimages \cite{comclip}. Hard-negative training can overfit the negative it was given \cite{hardpos}, and the deficit tracks the pretraining captions and the parallel encoder rather than any single architectural choice \cite{bindata,bindlimit}. A diagnostic result suggests a different route: binding information survives inside each single-modal embedding and is lost at cross-modal alignment \cite{bagcross,mate}, and enforcing region-to-segment alignment at inference recovers compositional accuracy without updating the encoders \cite{dualenc}. If the frozen features already carry the association structure the pooled score discards, the task is to read that structure rather than rebuild it beside the encoder.

\begin{figure*}[t]
\centering
\includegraphics[width=\textwidth]{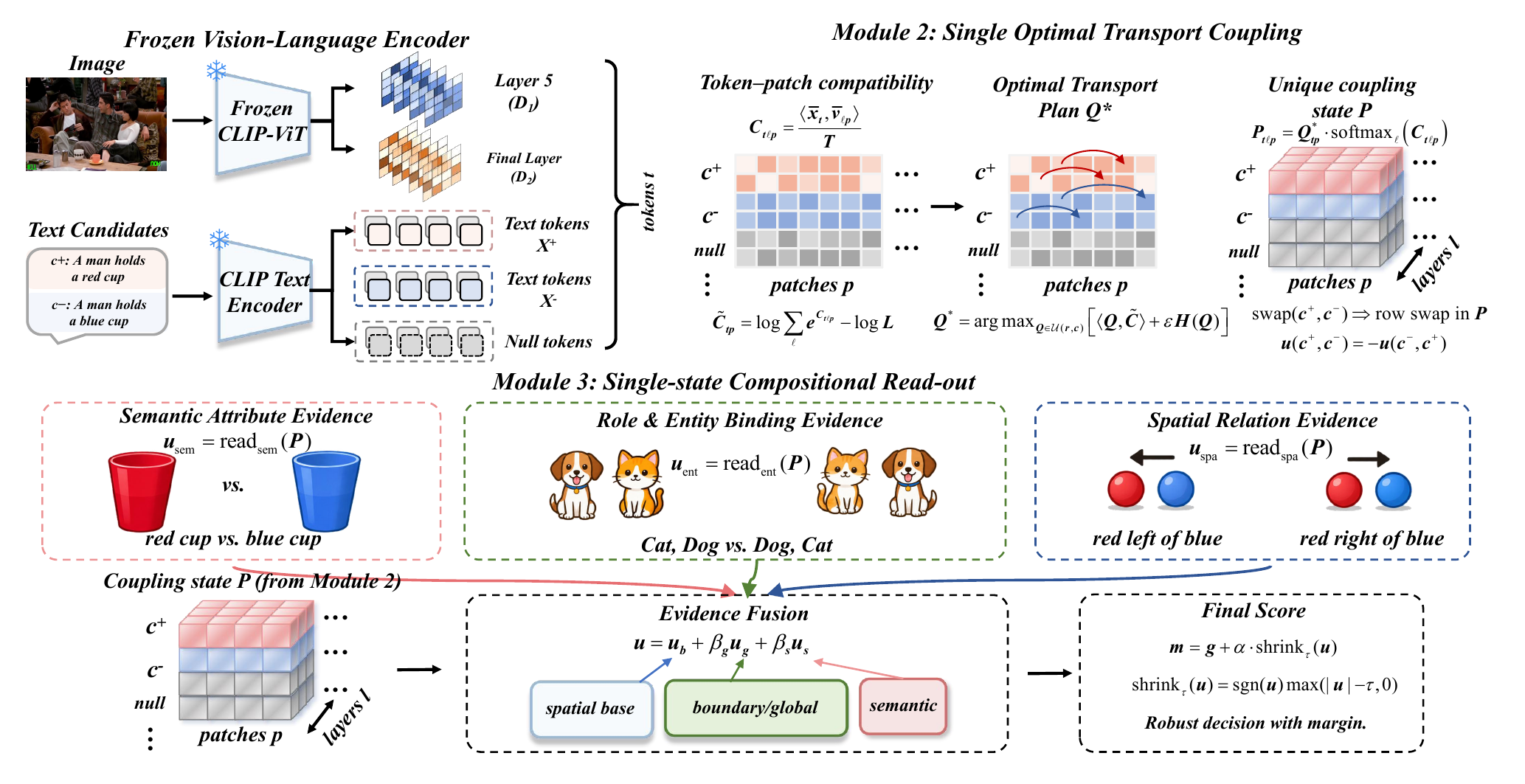}
\caption{Overview of BindCLIP. A frozen CLIP-ViT encodes the image and the candidate captions; a single token--patch--depth optimal-transport coupling $\mathbf{P}$ reads semantic, entity/role, and spatial evidence from one correspondence state; an antisymmetric pair score produces the preferred caption.}
\label{fig:framework}
\end{figure*}

Reading correspondence is what entropy-regularized optimal transport is for: matrix scaling produces a balanced plan with prescribed marginals and differentiates cheaply \cite{sinkhorn}, and in visual matching the same construction with explicit null states declines a match instead of inventing one \cite{superglue}; balancing likewise keeps patch selection from collapsing inside CLIP \cite{clippgs}. Unlike attention, which normalizes each query independently, two prescribed marginals force assignment to compete on both sides, the competition parallel encoders are reported to lack \cite{visstruct}. We also treat visual depth as part of the state rather than fixing one layer, since early and late transformer blocks retain different amounts of layout and semantics \cite{raghu}.

BindCLIP (Fig.~\ref{fig:framework}) is built on one latent object: a balanced token--patch--depth coupling shared by both candidate captions and several visual depths. Semantic, entity, order, and spatial evidence are read as energies of that single state, and exchanging the candidates permutes the plan so the score is exactly antisymmetric. A consensus refinement inside the same coupling contracts moves the candidates and the depths do not jointly support. Nothing is parsed, slotted, or re-wired. With locked parameters and a single inference path it improves the official What'sUp, ARO, and SugarCrepe metrics over frozen global CLIP, most strongly on the relation splits, and an inference-time lesion localizes spatial arrangement to the coupling.

\section{BindCLIP}
\subsection{One correspondence representation}
Let $X^+=\{x_i^+\}$ and $X^-=\{x_j^-\}$ be contextual tokens from two candidate captions, and let $V=\{v_{\ell p}\}$ denote patches at visual depth $\ell$ and location $p$. A shared projection maps text and vision to width $h$. We concatenate both token sets and append null token and null patch states. Their compatibility is
\begin{equation}
 C_{t\ell p}=\langle \bar x_t,\bar v_{\ell p}\rangle/T.
\end{equation}
After marginalizing depth, $\widetilde C_{tp}=\log\sum_\ell\exp C_{t\ell p}-\log L$, we compute
\begin{equation}
 Q^*=\arg\max_{Q\in\mathcal U(r,c)}\langle Q,\widetilde C\rangle+\epsilon H(Q),
 \label{eq:transport}
\end{equation}
with Sinkhorn scaling \cite{sinkhorn}, where $\mathcal U(r,c)$ is the polytope of nonnegative plans with row sums $r$ and column sums $c$.

\noindent\textbf{Why a balanced plan.} Cross-attention normalizes each token's distribution over regions independently: two tokens may claim the same region, and a region may be claimed without limit, so nothing in the operation forces the result to be a correspondence. Requiring both marginals instead makes the competition two-sided, since giving a patch to one token withdraws that capacity from the others, and this is what turns the plan from a saliency map into a binding.

\noindent\textbf{The marginals carry the comparison.} The row marginal $r$ divides the non-null mass equally between the two candidates, so neither can profit from being longer or from repeating a word. Within a candidate, mass is allocated by how much a token is not explained by the other caption: a soft cross-caption matching with a null option gives each token a residual against its best counterpart, and the row mass grows with that residual above a floor that keeps shared anchors present. The edited span, which is what distinguishes the pair, therefore receives the transport budget, while shared content stays available as context. The column marginal $c$ is image-dependent, obtained from each patch's accumulated token evidence, so uninformative regions are not forced to absorb mass. A fixed null mass on each side completes the polytope: a null patch absorbs tokens with no visual referent, and a null token absorbs patches that neither caption mentions.

\noindent\textbf{Depth as part of the state.} Solving Eq.~\eqref{eq:transport} on depth-marginalized evidence yields one plan rather than $L$ independent alignments; depth is then restored conditionally,
\begin{equation}
 P_{t\ell p}=Q^*_{tp}\,\operatorname{softmax}_{\ell}C_{t\ell p},
 \label{eq:joint}
\end{equation}
so a token--patch pair keeps a single transported mass and splits it across depths in proportion to where that pair is best explained, which lets one correspondence read layout and semantics without committing to either.

\subsection{Coupled semantic and spatial energy}
\label{sec:energy}
Each token carries a signed charge, positive for the first candidate and negative for the second, and $P$ transports that charge onto the patch grid, so the same plan yields a signed spatial field whose mass sits where the two captions disagree. The base readout takes studentized moments of the state
\begin{equation}
 \phi(v,p)=[v,p,p-a,v\otimes p,v\otimes(p-a)],
\end{equation}
where $a$ is the centroid of shared-token support. These terms answer complementary questions about one field: $v$ records what the disputed content is, $p$ where it lies in absolute image coordinates, and $p-a$ where it lies relative to the content both captions agree on, the frame in which ``left of'' or ``on'' is meaningful. The products $v\otimes p$, $v\otimes(p-a)$ record which content sits at that position, which is what separates an attribute swap from a mere co-occurrence. Reading them from one anchor is what makes the description relational. Layerwise studentization puts depths on a common scale, and attenuating moments on which the depths disagree yields a base score $u_b$.

Soft token mass on either side of the edited span aggregates two candidate support signatures $s^+_\ell$ and $s^-_\ell$ from the same $P$. Each signature concatenates four two-dimensional fields: centroid displacement, uncertainty-normalized displacement, and two smooth order statistics. Define
\begin{equation}
 s_{\Sigma\ell}=\tfrac12(s^+_\ell+s^-_\ell),\quad
 s_{\Delta\ell}=\tfrac12(s^+_\ell-s^-_\ell).
 \label{eq:support}
\end{equation}
An edit query interacts with shared support, an orthogonal context residual with candidate contrast, and their per-depth bilinear energy is
\begin{equation}
 E_\ell=\langle q_\Delta,s_{\Sigma\ell}\rangle+
 \langle q_\Sigma^\perp,s_{\Delta\ell}^\perp\rangle.
 \label{eq:factor}
\end{equation}
The two terms are deliberately crossed: the first asks whether the edit fits the layout the candidates share, the second whether the context the edit does not explain still distinguishes them. Projecting out the shared component prevents context from re-explaining the edit coordinate, so a caption cannot be preferred merely because its unedited words fit the image well. A separate studentized statistic compares semantic compatibility under the two candidate slices of $P$. All are energy readouts of the same state; the final score combines them,
\begin{equation}
 u=u_b+\beta_g u_g+\beta_s u_s,
 \label{eq:unified}
\end{equation}
with learned mixing weights $\beta_g,\beta_s$. The three are complementary statistics of one plan: $u_b$ says where the disputed mass lands, $u_g$ whether the disputed tokens are grounded rather than parked on the null state, and $u_s$ whether the matched regions are semantically compatible with the words assigned.
Exchanging the candidates permutes the coupling rows, and because the row marginal splits mass equally the permuted plan is the same plan with its candidate blocks swapped: shared quantities such as $s_{\Sigma\ell}$ and the anchor $a$ are invariant, signed quantities such as $s_{\Delta\ell}$ and the charge field change sign, so $u(X^+,X^-)=-u(X^-,X^+)$. The scorer has no preferred argument slot and cannot express an order prior, so a reported preference must come from the image.

\subsection{Consensus co-transport as a refinement}
The transport of shared support is a refinement that keeps a weakly supported move from perturbing the score. We build a barycentric signature $b_\ell$ from the mean candidate entity-support distributions. For each field $f$ let $\widehat s_{\Sigma\ell f}$ and $\widehat b_{\ell f}$ be unit directions. Candidate agreement and support concentration are
\begin{equation}
 a_{\ell f}=\tfrac12\!\left(1+\cos(s^+_{\ell f},s^-_{\ell f})\right),\quad
 r_\ell=(\kappa^+_\ell\kappa^-_\ell)^{1/4},
\end{equation}
where $\kappa^\pm_\ell\in[0,1]$ is the contraction of each entity-support distribution. The transport radius is $\eta_{\ell f}=a_{\ell f}^{4}r_\ell$ with harmonic cross-depth mean $\bar\eta_f$. The quartic power makes $\eta$ negligible unless the candidates almost coincide on that field, so partial agreement does not accumulate into a move; the fourth root and the harmonic mean are each dominated by their smallest argument, so a diffuse or dissenting depth suppresses the field for all of them. With global amplitude $q_0$ the shared field becomes
\begin{equation}
 \widetilde s_{\Sigma\ell f}=\|s_{\Sigma\ell f}\|\,\slerp(\widehat s_{\Sigma\ell f},\widehat b_{\ell f},q_0\bar\eta_f),
\end{equation}
and we apply the identical oriented rotation to $s_{\Delta\ell f}$, so the quotient changes coordinates of the complete representation rather than adding a parallel scorer ($q_0=0.85$). The image--caption margin combines the frozen global similarity with a small uncertainty dead zone,
\begin{equation}
 m=g+\alpha\shrink_\tau(u),\quad \shrink_\tau(u)=\operatorname{sgn}(u)\max(|u|-\tau,0),
 \label{eq:score}
\end{equation}
with global $(\tau,\alpha)=(0.005,0.225)$.

\subsection{Shared-structure objectives}
The learned coupling is constrained by caption preference, correct-versus-mismatched image interaction, structural sufficiency, region grounding when entity mapping is unambiguous, reflection consistency, and natural paraphrase consistency. Each loss uses smooth worst-group aggregation. The objectives stay distinct but all update the same $P$ and the same score, so an improvement on one has to survive the others. Consensus co-transport introduces no additional trainable parameters. The scorer has 521,475 in total, and its learned components are the shared projections and the relation and semantic scale logits.

For $N$ tokens, $R$ patches, $L$ visual depths, width $h$, and $K$ Sinkhorn iterations, compatibility and scaling require $O(NRLh+KNR)$ work and $O(NRL)$ coupling memory. Co-transport is linear in the stored support fields and starts no second transport solve.

\begin{figure*}[t]
\centering
\includegraphics[width=\textwidth]{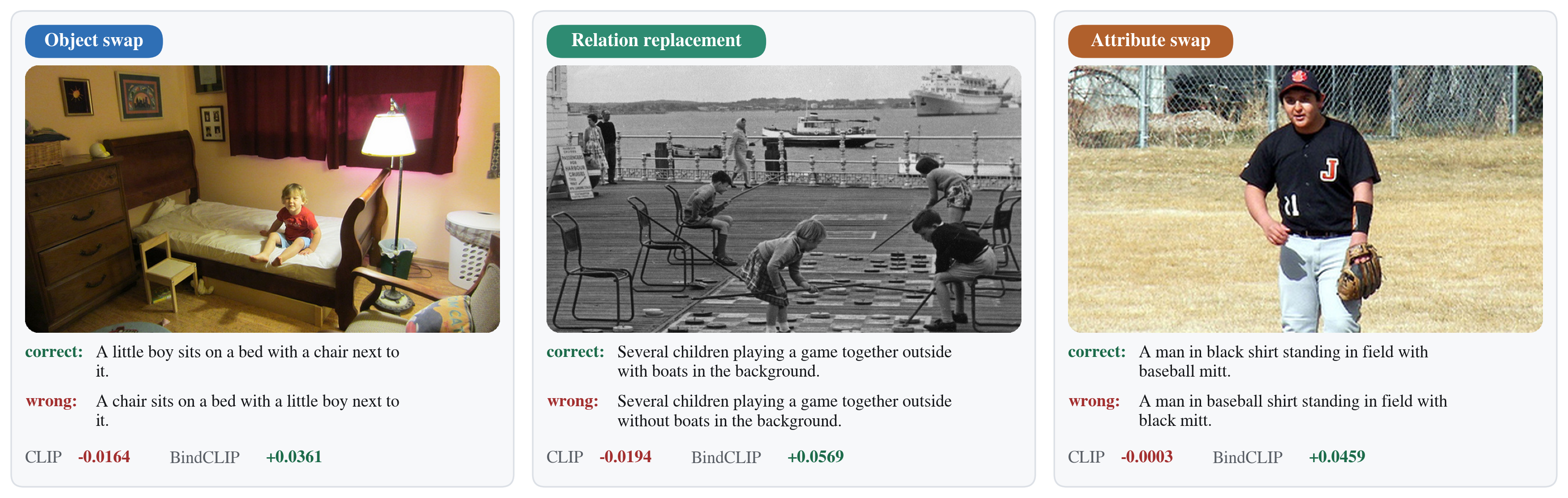}
\caption{Three examples from the official SugarCrepe split, one per phenomenon, where frozen global CLIP prefers the compositionally wrong caption (negative margin) and BindCLIP reverses it (positive margin). Columns: object swap, relation replacement, attribute swap. In the object- and attribute-swap columns the two captions contain identical content words and differ only in which word binds to which region.}
\label{fig:qual}
\end{figure*}

\section{Experiments}
\subsection{Data and protocol}
We froze CLIP ViT-L/14-336 and used its layer-5 and final $24\times24$ patch grids. Training and development used 143,188 and 30,085 image-disjoint COCO pairs across eight related families: entity and attribute replacement, role swap, negation, relative geometry, absolute frame position, and natural depth and vertical relations. The projected width was 192, captions were truncated to 32 tokens, and coupling used temperature 0.1, 64 Sinkhorn iterations, visual null mass 0.50, text null mass 0.25, and uncertainty floor 0.05.

ARO image paths were deterministically split into train (hash buckets 0--2), validation (3--4), and holdout (5--9). The quartic contraction and $q_0$ were fixed using ARO train and COCO development; What'sUp and ARO validation/holdout were then evaluated once for the locked candidate. Training used AdamW with weight decay $10^{-3}$ at batch size 64 for three epochs of 100 steps, with learning rate $10^{-4}$ on the newly introduced parameters and the inherited transport parameters held fixed; development selected the step-200 checkpoint.

\begin{table}[t]
\caption{Main results. Top: the official headline metric of each benchmark, frozen CLIP ViT-L/14-336 vs.\ BindCLIP reading the same frozen features. Bottom: the binding-specific slices our mechanism targets. Intervals are paired bootstrap over items, $10^4$ resamples. $^\ddagger$averaged over the two controlled suites; pair accuracy is scored over the 410 arrangement pairs, not over the 820 individual items.}
\label{tab:main}
\centering
\setlength{\tabcolsep}{2.5pt}
\resizebox{\columnwidth}{!}{%
\begin{tabular}{llrrrr}
\toprule
Benchmark & Official metric & $n$ & CLIP & BindCLIP & $\Delta$ (95\% CI) \\
\midrule
\multicolumn{6}{l}{\emph{Overall (official headline metric)}} \\
ARO & Macro over splits & 52,685 & 57.69 & \textbf{60.33} & $+2.63\;[+2.24,+3.04]$ \\
What'sUp$^\ddagger$ & Individual & 820 & 27.68 & \textbf{29.13} & $+1.45\;[-1.71,+4.76]$ \\
SugarCrepe & Item acc. & 7,511 & 78.38 & \textbf{79.23} & $+0.85\;[-0.02,+1.73]$ \\
\midrule
\multicolumn{6}{l}{\emph{Relational and attribute-binding slices}} \\
ARO & Relation & 23,937 & 53.75 & \textbf{57.19} & $+3.44\;[+2.86,+4.06]$ \\
ARO & Attribution & 28,748 & 61.64 & \textbf{63.46} & $+1.82\;[+1.29,+2.35]$ \\
What'sUp$^\ddagger$ & Pair, strict & 410 & 0.98 & \textbf{4.14} & $+3.17\;[+1.22,+5.12]$ \\
SugarCrepe & Replace-rel. & 1,406 & 66.86 & \textbf{71.41} & $+4.55\;[+2.13,+6.90]$ \\
SugarCrepe & Swap-attr. & 666 & 62.31 & \textbf{65.47} & $+3.15\;[-0.45,+6.61]$ \\
\bottomrule
\end{tabular}%
}
\end{table}

\subsection{Main results}
\noindent\textbf{Evaluation protocol.} Table~\ref{tab:main} is scored with the benchmarks' own public metrics from one locked scorer that sees no benchmark identity and no relation name. Prior compositional repairs either update the encoder weights or add a parser, a relation inventory, or an external detector, so a difference against them would confound the mechanism with the representation it reads; we therefore hold the backbone fixed and compare against frozen global CLIP reading identical features.

All three official headline metrics improve, and What'sUp shows why its strict criterion is the informative one. The pair criterion, which credits a pair only when both arrangements of the same objects are matched, rises from 4 of 410 pairs to 17 at an exact McNemar $p=0.002$, while individual accuracy over the same items cannot register the difference. A fixed preposition preference cannot satisfy it: the two images of a pair differ only in how the same objects are arranged, so answering both the same way is wrong on one of them by construction. The ARO relation split makes the same point at scale, and the lesion reported below damages that split and no other benchmark split. The development diagnostic shows the same shape rather than an even lift: entity replacement, where pooled similarity is already high, moves least, while the families whose pooled baseline sits nearest chance gain up to $+21.9$.

SugarCrepe separates the same two regimes at the category level. Replacing a relation or swapping two attributes leaves the bag of content words almost or exactly unchanged, so pooled similarity barely separates them and the decision has to come from correspondence; these are the categories that move. Categories that replace or add an object change what is present instead, which a pooled embedding already measures well, so the seven-category aggregate averages two regimes rather than measuring one effect; we take the category structure, not the total, as what the mechanism predicts. Three independent trainings land within $0.03$ points of each other and all above frozen CLIP.

\noindent\textbf{Patch access is not the explanation.} Once a scorer reads patch-level features instead of one pooled embedding, the improvement might follow from the features rather than from the coupling. We test this with zero-parameter local-alignment baselines over exactly the tensors BindCLIP consumes: untrained cosines between each projected caption token and each patch feature, pooled by max or by temperature-softmax. Every variant scores below frozen global CLIP on SugarCrepe, because pooling a token--patch similarity is a saliency statistic: it reports how strongly some region matches some word and lets both captions claim the same region without cost, so it measures presence and not assignment.

Adding a local margin to the global one rather than replacing it does exceed the frozen baseline, at four of six pooling choices and most weights, and the best combination gains $1.81$ points on the seven-category total against BindCLIP's $0.85$. Its advantage is linguistic rather than compositional. SugarCrepe forms the add-attribute and add-object negatives by appending a phrase, so the negative is systematically longer and preferring the shorter caption is nearly free accuracy; only 35 of those pairs have captions of equal length at all. Restricting to the length-matched pairs removes that information and the additive advantage disappears: no pooling choice at any weight is then distinguishable from frozen CLIP, and the combination selected on the full set falls to $-0.73$. The totals settle nothing on that subset; the categories do. On the two that turn on which word binds to which region the split is outright: the additive baseline falls to $-4.07$ on relation replacement where BindCLIP reaches $+3.39$, and attribute exchange separates the same way. The baseline wins where length decides and loses where binding decides.

\noindent\textbf{Qualitative.} Fig.~\ref{fig:qual} shows the same correction case by case on three official SugarCrepe pairs. The two exchange cases are the sharpest test available: both captions hold identical content words, so no lexical, length, or presence cue can separate them.

\subsection{Mechanism and reliability analysis}
To ask what the plan itself contributes, we hold the locked parameters fixed and replace only the plan at inference with the outer product of its own marginals. Every token keeps the mass it was given and every patch the mass it received, and the same readouts run on top, so the only quantity removed is which token binds to which patch. The cost is large and selective. Accuracy on the development families falls by $9.49$ points, and the eight separate without overlap: the four spatial ones lose between $4.7$ and $14.3$ points each, while entity replacement, attribute replacement, role swap, and negation lose at most $2.6$. The dissociation repeats on ARO with opposite signs: the lesion costs $0.97$ points on the relation split and returns $0.88$ on attribution. Relation is thus at once the benchmark split with the largest gain and the only one the lesion takes back. Attribution asks which adjective belongs to which noun, which the semantic and entity readouts answer from what is present; relation asks how two regions are arranged, which has no answer unless the plan states where each region is. The coupling is therefore what carries arrangement.

\end{document}